\documentclass[letterpaper, 10 pt, conference]{ieeeconf}
\usepackage[utf8]{inputenc}
\usepackage{bm} 
\usepackage{booktabs}
 \usepackage{balance}

\IEEEoverridecommandlockouts                              

\usepackage{graphicx} 

\title{\LARGE \bf
Surfing on an uncertain edge: Precision cutting of soft tissue\\ using torque-based medium classification
}

\author{Artūras Straižys, Michael Burke and Subramanian Ramamoorthy
\thanks{Authors are with School of Informatics, University of Edinburgh}%
}

\begin{document}

\maketitle
\thispagestyle{empty}
\pagestyle{empty}

\begin{abstract}

Precision cutting of soft-tissue remains a challenging problem in robotics, due to the complex and unpredictable mechanical behaviour of tissue under manipulation. Here, we consider the challenge of cutting along the boundary between two soft mediums, a problem that is made extremely difficult due to visibility constraints, which means that the precise location of the cutting trajectory is typically unknown. This paper introduces a novel strategy to address this task, using a binary medium classifier trained using joint torque measurements, and a closed loop control law that relies on an error signal compactly encoded in the decision boundary of the classifier. We illustrate this on a grapefruit cutting task, successfully modulating a nominal trajectory fit using dynamic movement primitives to follow the boundary between grapefruit pulp and peel using torque based medium classification. Results show that this control strategy is successful in 72 \% of attempts in contrast to control using a nominal trajectory, which only succeeds in 50 \% of attempts.

\end{abstract}

\section{INTRODUCTION}

The design of effective learning and adaptive control strategies for precision cutting remains an open problem in robotics \cite{Long,vegetables,mechcut}. This is particularly true for the case where cutting involves moving between two mediums, and where there is uncertainty in the location of these.

Cutting tasks of this form are regularly encountered in surgery, where tumour extraction is guided by well established continuum differences between tumours and normal tissue \cite{cristini2010multiscale,mcknight2002mr}.
This paper is motivated by wide local excision, a surgical procedure that aims to remove a tumour with a clear margin of healthy tissue around it. At present, manipulation tasks like these are inconceivable for autonomous robots, for a variety of reasons. First, the constrained operational space of non-trivial geometry restricts an end-effector's maneuverability and is severely limited by visibility constraints. These visibility constraints are a particular challenge, and human surgeons often rely strongly on haptic feedback for cutting, using tactile tissue differences to guide procedures instead of vision. In addition, this task is highly variable and uncertain, due to the unpredictable behaviour of deformable tissue and varied tumour shapes. Finally, and most importantly, this contact-rich task is characterised by safety constraints imposed on the region of operation and allowable applied forces. It is therefore critical to keep an end-effector inside a desired region while executing the task.

In this study, we move towards addressing the challenge of autonomous cutting with visibility constraints by 1) employing probabilistic inference to identify the boundary between two mediums using torque sensing, and 2) using the medium classification probability as an error signal for online, closed-loop movement adaptation. As a testbed, we consider fruit processing, and study the task of scooping a grapefruit segment out of the membrane with a kitchen knife (see Fig. 1). This manipulation task shares several important characteristics with the surgery problem described above, including the complex geometry of the task space, the need for contact-rich manipulation in a deformable environment and the presence of two mediums with differing material properties. Precision food processing is itself an industrially useful skill, and the ability to extract fruit portions without damaging food products is particularly valuable. 

\begin{figure}[t]
    \centering
    \includegraphics[scale=0.085]{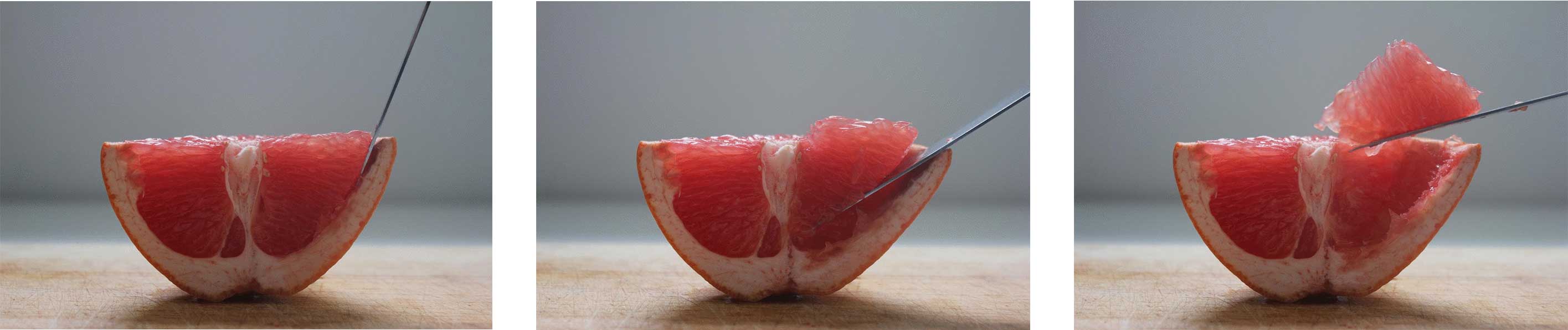}
    \caption{Scooping grapefruit with a regular paring knife.}
\end{figure}

A key feature of the grapefruit testbed is an implicit task requirement to keep the knife inside the intermediate region between the peel and pulp boundary, such that most of the pulpy segment is extracted without the knife getting stuck in the peel, or too much grapefruit being left on the membrane. Since the exact shape and location of this boundary curve is unknown, it must be inferred during task execution. The key insight for the approach proposed in this work comes from observing grapefruit scooping when executed by human. It is clear that humans do not rely on an accurate geometrical model of the fruit, but instead apply a general scooping movement that is continued until \textit{it stops feeling right} (i.e. when the knife starts progressively entering the peel). In these cases, the movement is either adjusted, or completely restarted using a different insertion angle. Our hypothesis is thus that cutting occurs using a rough nominal trajectory that is modulated by torque feedback resulting from the differing tissue characteristics of the two mediums being separated. 

This work introduces a novel framework to accomplish tasks of this form. Here, we use Dynamic Movement Primitives (DMP) to learn a general scooping motion (the nominal trajectory) using kinesthetic demonstration. However, due to the variations in grapefruit's shape and its mechanical properties, we show that generalisation of the learned movement primitive is inadequate. We therefore propose a control scheme in which the corrections to the DMP trajectory reflect the probability of knife being inserted into either of the two mediums. In this formulation, the point of highest uncertainty in this belief (probability of 0.5) serves as a proxy for the desired region of operation (i.e. the boundary between the pulp and peel). This probability is estimated at each time step of task execution by classifying torque readings from the joints of robot arm. We use a logistic regression classifier trained to disambiguate the mediums on a dataset of multiple task executions to demonstrate the feasibility of this method. 

\section{RELATED WORK}

There is a substantial amount of research on the use of the force feedback in robotic manipulation tasks. However, most work is focused on rigid object manipulation (e.g. there is extensive research on the use of force data in the areas of robot door opening \cite{door1, door_grasp, door3}, grasping \cite{door_grasp, Pastor, grasp} and object identification \cite{obj_ident1, obj_ident2}), where task dynamics are relatively well understood and many mature techniques for motion planning and control are readily available. Unfortunately, many of these techniques are not applicable to the manipulation of deformable objects.

A particularly representative deformable object manipulation task involves food cutting, for example fruits or cheese. This process has time-varying nonlinear dynamics that are extremely difficult to describe analytically \cite{nonlinearcut} (although attempts have been made \cite{mechcut,adaptcut,salami}). As a result, learning-based techniques have been proposed to address this challenge. Lenz et. al \cite{DeepMPC} use deep learning techniques to model food cutting tasks and further use these in a model-predictive control scheme. Here, robot controls are optimised in real time with respect to the constructed cost function, which penalises the height of the knife and its deviation from a sawing range, thus forcing a cutting movement. This approach was verified on a variety of food objects, such as lemons and potatoes, and showed its ability to adapt to both intra-class and time-varying variations in the physical properties of the objects. 

A similar approach to learning the predictive model is described by Tian et. al \cite{tactileMPC}, where the authors demonstrated tactile servoing using high dimensional tactile sensor data. Another use of learning in the latent space is presented in the work of Gemici and Saxena \cite{pr2salad}, which is concerned with robotic handling of food objects, e.g. grasping or piercing. Here, latent features of objects were learned from force data collected during the manipulation and then used to classify the objects for manipulation planning. 

Many manipulation tasks, e.g. scooping, involve nontrivial kinematic trajectories that can be learned from demonstration. In \cite{Ijspeert}, the authors propose a general framework (Dynamic Movement Primitives, or DMPs) for encoding complex movements as a parameterised policy. This framework, when coupled with feedback enables reactive movement adaptations  \cite{DMPobstacle, Pastor}. In our work, we use a predictive model of the expected sensor trace, that is based on the statistics of multiple task executions. This approach is similar to \cite{outcome}, where the statistics of sensor measurements from the successful task executions were used to construct a predictive model for online failure detection. We apply a similar method to model the region of operation (pulp or peel in our grapefruit example) using torque sensor readings. However, a key contribution of this work is the use of this classification scheme to perform boundary identification for cutting with visibility constraints, through the introduction of a control scheme for movement adaptation based on the estimated probability of being in a given medium.

\section{PROBLEM FORMULATION}
As discussed previously, this paper focuses on the task of precision-cutting between two mediums. Our primary interest lies in the development of an adaptive control framework, so we do not consider the use of any task specific cutting equipment or machinery, and focus on cutting using a standard kitchen knife. 
In addition, we allow control of the initial insertion position, and thus, we manually initialise the starting position of the knife. 

\begin{figure}[t]
    \centering
    \includegraphics[scale=1.5]{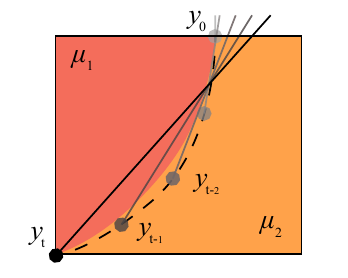}
    \caption{Medium separation by following nominal trajectory. $\mu_1$ and $\mu_2$ are stiffness parameters of mediums, $y_0$ is initial pose and $y_t$ is trajectory of the knife. The dashed line represents the nominal path for tip of the knife.}
\end{figure}

The task described above can be formulated in a 2D task space. Consider two elastic mediums with different stiffness ($\mu_{1}$ and $\mu_{2}$), separated by a curved boundary (see Fig. 2). Assume a strong prior over the stiffness and boundary curve (dashed line) is available, but no exact parameters are known. The objective is to steer the tip of the knife along the true boundary such that separation of mediums is maximised, while avoiding excessive deformations imposed to either of mediums by the knife. Since the exact curve of the boundary is unknown, the open loop execution of the prescribed path (based on a prior belief over the curve of the boundary) runs the risk of inserting the knife into the peel (in our grapefruit example), thereby severely restricting the knife's maneuverability. 

\section{Cutting using uncertainty feedback}

We address the challenge above using a learning strategy where the desired operational region is compactly encoded in the decision boundary of a binary medium classifier. Here, the estimated likelihood of sensor readings associated with either medium guides the movement execution in the form of online trajectory correction. In summary, we  1) use the DMP framework to encode a nominal scooping trajectory, 2) learn probabilistic classification of sensor readings associated with operation in either mediums, and 3) construct a control scheme that corrects the DMP according to the estimated posterior distribution over either medium, as illustrated in Fig. 3. 

\begin{figure}[t]
    \includegraphics[width=0.48\textwidth]{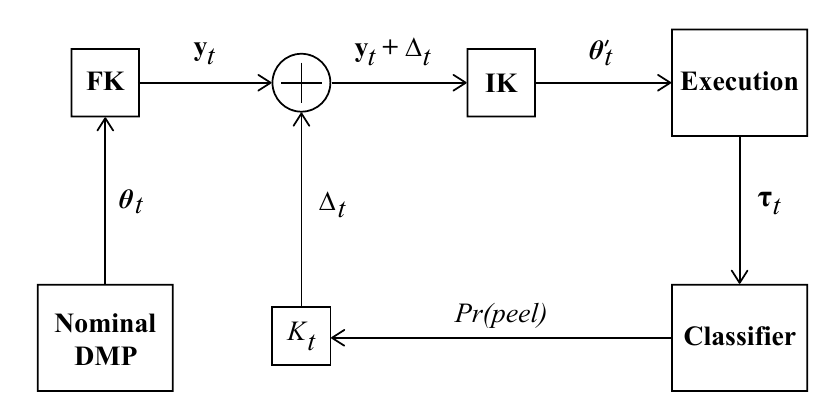}
    \caption{Overview of the proposed control scheme.}
\end{figure}

Here, $\bm{\theta}_t$ denotes the nominal joint trajectory, $\bm{\theta}^{'}_t$ the corrected joint trajectory, and $\mathbf{y}_t$ represents the nominal Cartesian trajectory. $\Delta_t$ denotes the correction term applied to the Cartesian trajectory, $K_t$ is a gain matrix, $\bm{\tau}_t$ refers to the sensed torque readings at time step $t$, and $Pr(peel)$ is the probability of the knife being inserted into one of the mediums. FK and IK are Forward and Inverse Kinematics transformations for the robot arm. Note that gain matrix $K_t$ is time dependent, as the correction direction depends on the position along the nominal trajectory at time $t$. We briefly discuss each element in the control framework below.

\subsection{Nominal trajectory modelling using DMPs}
In the DMP formulation \cite{Ijspeert}, any goal-oriented movement primitive can be expressed as:
\begin{equation}
    \tau\ddot{y} = \alpha_{z}(\beta_{z}(g - y)-\dot{y}) + f
\end{equation}
where $\ddot{y}$, $\dot{y}$ and $y$ are the desired acceleration, velocity and position, respectively, $g$ is the goal position, $\tau$ is a temporal scaling factor, $\alpha_{z}$ and $\beta_{z}$ are time constants, and $f$ is a nonlinear forcing function. In the above equation, the nonlinear term $f$ modulates the landscape of a global point attractor $g$. Thus, an arbitrarily complex movement can be represented by appropriately constructing $f$. Typically, the nonlinear function $f$ is represented using a normalized linear combination of basis functions:
\begin{equation}
    f(x) = {\frac{{\sum_{i=1}^{N} \Psi_{i}}(x)w_{i}}{\sum_{i=1}^{N} \Psi_{i}}(x)}x(g-y_{0})
\end{equation}
where $N$ is the number of basis functions $\Psi_{i}(x) =e^{-h_{i}(x-c_{i})^2}$ with center $c_{i}$, widths $h_{i}$ and weights $w_{i}$. 

Note that the forcing term does not depend on time, but does depend on phase variable $x$ that monotonically decays from 1 to 0 with a user specified rate $\alpha_{x}$:
\begin{equation}
    \tau \dot{x}=-\alpha_{x}x.
\end{equation}

In the proposed framework, a nominal scooping trajectory is captured by kinesthetically guiding the robot arm and recording the end-effector’s Cartesian path (the time series of the end-effector’s position and orientation waypoints). The corresponding velocity and acceleration profiles of the movement, $\dot{y}$ and $\ddot{y}$, are obtained by twice differentiating the recorded end-effector's path $y$. The desired nonlinear function $f$ (expressed by rearranging (1)) is approximated by employing the Locally Weighted Regression method \cite{lwpr}, used for optimising the weights of the basis functions. It should be noted that the nominal trajectory could be modelled using any behaviour cloning strategy, and the proposed approach is not limited to the use of DMPs. 

In our approach, at each time step $t$ we add a local correction $\Delta_t$ to the current point on the Cartesian path $y_t$ of nominal DMP (see Fig. 3). The correction term $\Delta_t$ is given by

\begin{equation}
    \Delta_t = K_t\Big[Pr(m,t) - 0.5\Big]
\end{equation}
where $K_t$ is time-varying positive definite gain matrix that defines the sensitivity of task variables and $Pr(m,t) \in [0,1]$ is the probability of the knife being in the medium $m$ at time step $t$. Note, that the desired region of operation at each time step $t$ lies at the boundary between two mediums, where probability $Pr(m)$ is equal to 0.5.

Thus, our proposed uncertainty driven control law can be formulated generally as

\begin{equation}
    y{'}_t = y_t + K_t\Big[Pr(m,t) - 0.5\Big]
\end{equation}
where $y{'}_t$ is the corrected version of the nominal trajectory $y_t$.

\subsection{Logistic regression}

In this work, we use logistic regression to model the probability of being in a given medium. In this approach, model parameters $w$ are fit by maximizing the probability of the data under a linear logistic model:

\begin{equation}
\mathcal{L}(w) = \prod_{i = 1}^{N} p(y_i|x_i,w)
\end{equation}

where $\mathcal{L}$ is the likelihood, $N$ is the number of training samples of torque readings, $y_i$ is the label (e.g. ``Peel'' or ``Pulp'') of the $i$ th example of torque data, $x_i$ is a vector of torque readings of the $i$ th example and $w$ is a model parameter.

If the cost function $J(w)$ is defined as the negative log-likelihood of labels $y$, then the above expression is equivalent to minimizing:

\begin{equation}
    J(w) = \sum_{i=1}^{N} \Big[-y_i \ log(\sigma(w^T x_i)) - (1 - y_i) \ log(1 - \sigma(w^T x_i) \Big]
\end{equation}

where $\sigma(\cdot)$ is a sigmoid function and labels $y \in \small\{0,1\small\}$.

In order to discourage the optimizer from overfitting to the training data, the cost function can include an additional regularization term that penalizes extreme weight coefficients, e.g. $ {\lambda \over{2}} ||w||^2$, where $\lambda$ denotes the regularization strength.

\begin{figure*}[ht]
    \centering
    \includegraphics[width=\textwidth]{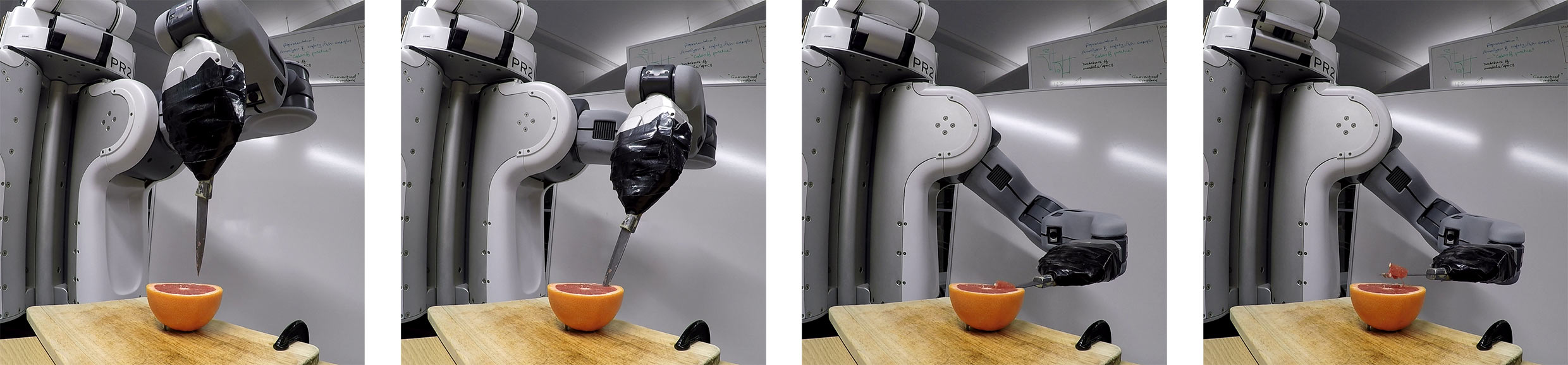}
    \caption{Images of the PR2 robot scooping a grapefruit.\label{fig:pr2}}
\end{figure*}

\subsection{Experimental setup}

All experiments were conducted using a 7 degree-of-freedom PR2 robot arm. The PR2 arm is counterbalanced and highly compliant, and is well-suited for kinesthetic demonstrations of flexible and fluid movements. The remaining elements of the experimental setup consisted of a chopping board clamped to the table, a halved grapefruit fixed to the chopping board, and a regular paring knife secured at the gripper (see Fig. \ref{fig:pr2}). For registering the torques experienced at the joints of the arm we used standard PR2 joint effort readings (a joint torque estimate based on the joint motor current).

\subsection{Evaluation of nominal DMP}

The learned scooping DMP was evaluated on the 10 randomly chosen segments. Before each trial, a segment was pre-cut along the segment radii and the pose of the knife was manually adjusted, as discussed in the previous section. Successful task execution implies the complete extraction of the undamaged segment without the knife getting stuck in the peel.

The results of the trials agreed with original expectations, with only 2 successful task executions out of total 10. In 7 of the failed trials, the knife entered the peel and the execution was aborted. Moreover, an instance of tearing apart the segment during the scooping was registered. As anticipated, the main difficulty of the task was avoiding the knife's insertion into the peel, where further knife maneuverability became limited. 

\section{LEARNING THE BOUNDARY REGION USING SENSED TORQUE}

\subsection{Dataset}

The learned DMP was used to accumulate joint torque readings associated with successful (cut through the flesh) and failed task executions (cut into the peel). These traces of torque measurements were analysed and further used for training the logistic regression model to estimate the probability of the knife's deviation from the desired region of operation, i.e. the boundary between pulp and peel, where task executions succeed. We used the learned DMP and experimental setup described in the previous section. The criteria for a successful trial remained unchanged from the preliminary evaluation of the DMP. A total of 111 scooping trials were conducted using a number of grapefruit, of which 55 trials were successful and 56 trials failed.

\begin{figure*}[t]
    \centering
    \includegraphics[width=\textwidth]{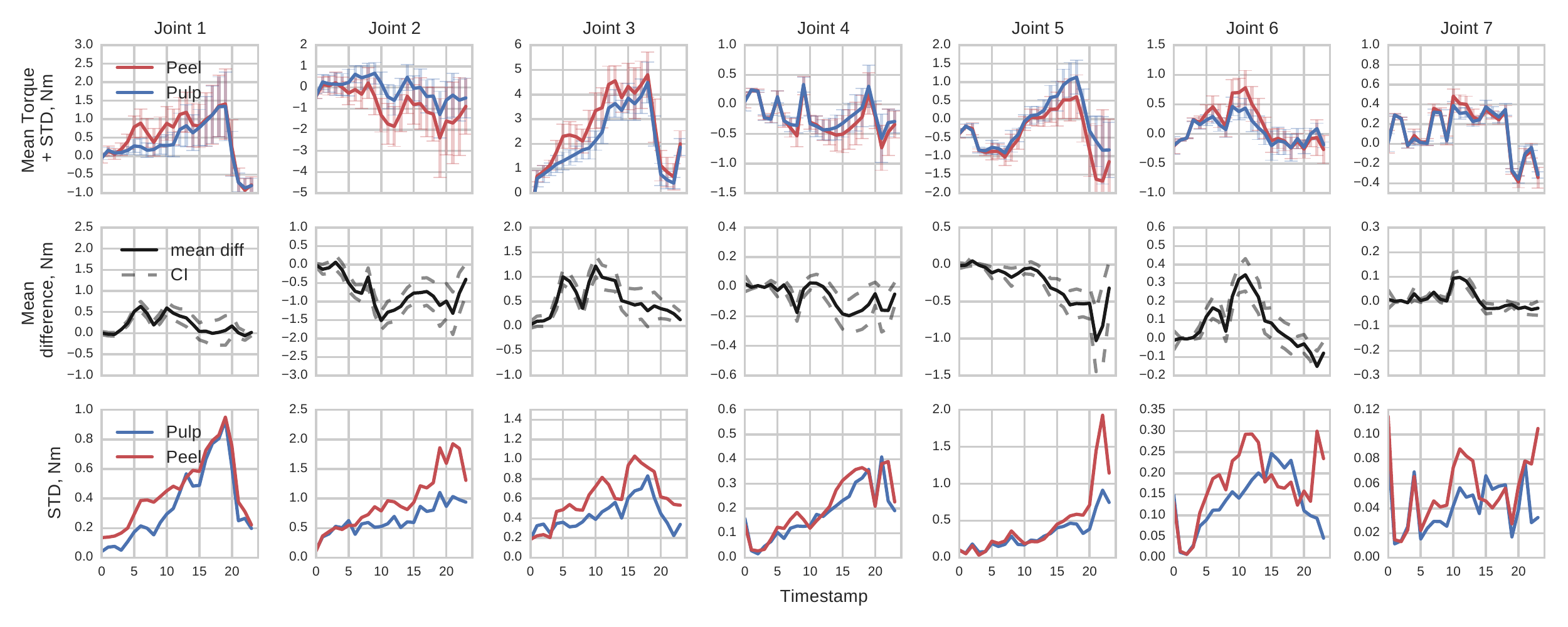}
    \caption{Summary statistics for collected torque data. Note that the most dominant joints in the used DMP are joints 1, 2 and 3. \label{fig:summary_stats}}
\end{figure*}

The nominal trajectory comprised 24 segments, at which a single snapshot of torque readings was taken. Thus the recorded data consisted of 24 time-indexed 7-dimensional vectors. Fig. \ref{fig:summary_stats} shows the descriptive statistics of the collected data. 

\subsection{Classification}

The dataset of 111 trials was randomized and split into 90 sensor traces allocated for training and validation and 21 traces held out for testing. The training and validation dataset consisted of 44 examples of ``Pulp'' torque traces and 46 examples of ``Peel''. As discussed, each trace contained of 24 time-indexed samples of torque reading for each of the 7 joints. Thus, in total the training and validation dataset contained 1,056 and 1,104 individual examples of ``Pulp'' and ``Peel'' torque readings, respectively. The classifier's input comprised of an 8-dimensional vector (7 torque readings for each of the joints, plus the time index). The objective of the classification task was to estimate the probability of the measurement being taken inside of either medium given the current torque measurements. In our approach, a desirable property of a classifier is to be robust to the outliers and to handle the ambiguous inputs by reporting the appropriate levels of uncertainty (i.e. to avoid being overconfident). We used a logistic regression model, which we validated using the K-fold cross-validation technique with 10 folds. Thus, each fold used 81 examples for training and 9 examples for validation. The validation and test results are given in Table I. 

\begin{figure}
    \centering
    \includegraphics[width=0.48\textwidth]{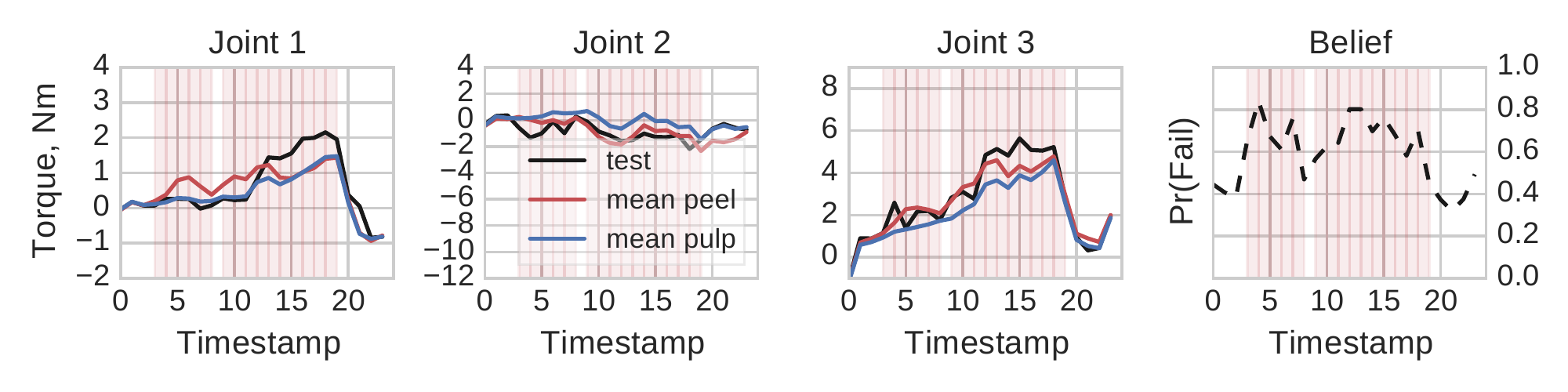}
    \caption{Medium classification based on torque readings. The test trace (black solid line) represents a trial where the knife got stuck inside the peel at the middle of the execution until its final release at time step $t=19$. The red background signifies the classifier's binary decision (red denotes the peel) at time step $t$. Note, that only 3 most dominant joints are displayed.\label{fig:example_traj}}
\end{figure}

It should be noted that the modelling approach has an inherently noisy training process. Since all of the 24 torque samples in the sensor trace share the same label defined by the task outcome, all the intermediate phenomena are disregarded. For instance, if knife was closely following the desired boundary region throughout most of the execution but got stuck in the peel at the very end, all of the 24 torque readings would be labeled as ``Peel''. However, the ability to capture the uncertainty demonstrated by Logistic Regression model alleviates this issue (see Fig. \ref{fig:example_traj}), and in some respects this training process forces a more conservative probabilistic model. It is important to note that despite misclassifying some intermediate samples, the trained model does not commit to any extreme beliefs over the mediums, unless the test input is strongly representative of a given class. Finally, in the case of ambiguous test inputs (i.e. where torque levels of the input trace appear uncharacteristic for a given label), the model demonstrates desirable levels of uncertainty, which is extremely important given the fact that we seek to use this for feedback control.

\section{ONLINE DMP ADAPTATION}

\begin{table}
\caption{Logistic regression model evaluation}
\label{tab:my-table1}
\begin{tabular}{r|c|c||c|c|}
\cline{2-5}
\multicolumn{1}{l|}{} & \multicolumn{2}{c||}{\textbf{Validation}} & \multicolumn{2}{c|}{\textbf{Test}} \\ \cline{2-5} 
\multicolumn{1}{l|}{} & \begin{tabular}[c]{@{}c@{}}Predicted \\ Peel\end{tabular} & \begin{tabular}[c]{@{}c@{}}Predicted \\ Pulp\end{tabular} & \begin{tabular}[c]{@{}c@{}}Predicted\\ Peel\end{tabular} & \begin{tabular}[c]{@{}c@{}}Predicted\\ Pulp\end{tabular} \\ \hline
\multicolumn{1}{|r|}{Actual Peel} & 673 & 431 & 174 & 66 \\ \hline
\multicolumn{1}{|r|}{Actual Pulp} & 306 & 750 & 65 & 199 \\ \hline
\multicolumn{1}{|r|}{\textit{Sensitivity}} & \multicolumn{2}{c||}{0.61} & \multicolumn{2}{c|}{0.73} \\ \hline
\multicolumn{1}{|r|}{\textit{Specificity}} & \multicolumn{2}{c||}{0.71} & \multicolumn{2}{c|}{0.75} \\ \hline
\multicolumn{1}{|r|}{\textit{\begin{tabular}[c]{@{}r@{}}Misclas.\\ rate\end{tabular}}} & \multicolumn{2}{c||}{34\%} & \multicolumn{2}{c|}{26\%} \\ \hline
\end{tabular}
\end{table}

We reused the nominal DMP and trained classifier from the previous section and constructed the closed-loop control scheme as shown in Fig. 3. First, the nominal joint trajectory associated with learned DMP is transformed into the end-effector's trajectory in the Cartesian space, where all the required corrections are relatively straightforward. As discussed previously, we use the probability of the knife being inserted into the peel for deriving the required motion corrections. We used a simple motion correction scheme for experimentation, in which the first half of the scooping motion (where the most dominant movement component involves pushing the knife downwards) is modulated towards the center of the grapefruit. In the second half of the movement (where the knife slides under the segment while moving towards the center of the grapefruit), the motion was modulated upwards. Thus, in each of the cases, the knife deviates from the peel region towards the pulp region, when the estimated probability of peel increases. For both cases we used a gain of 0.01m (i.e. 100$\%$ probability of peel would translate the movement 10mm away from the nominal trajectory in the prescribed direction). It should be noted that more complex schemes can be applied, e.g. modulation in the direction of the normal to the side of the knife.

\begin{table}
\centering
\caption{Method comparison}
\label{tab:my-table2}
\begin{tabular}{@{}rcc@{}}
\toprule
\multicolumn{1}{l}{Control scheme} & \textbf{Open-loop} & \textbf{Closed-loop} \\ \midrule
\textit{Successful trials} & 55 & 36 \\
\textit{Failed trials} & 56 & 14 \\
\textit{Success rate} & 50\% & 72\% \\ \bottomrule
\end{tabular}
\end{table}

For this experiment we conducted 50 trials of grapefruit scooping on randomly chosen segments in a total of 12 different grapefruits. As in the previous section, a successful trial required the complete extraction of an intact segment without the knife getting stuck inside the peel. The results are provided in the Table II. 36 out of 50 trials achieved successful task completion. In all of the 14 failed attempts, the knife entered the peel at the start of the cut and propagated deeply before the peel could be classified. In these cases, the movement corrections towards the center of the grapefruit failed, as the knife could not tear the peel with the side of the blade. Perhaps, in such cases the DMP can be reversed and re-applied with estimated corrections, as the classifier successfully reflected the event of knife being stuck in the peel. 

In the successful trials, the knife visibly responded to the local increase in the resistance throughout the movement execution. It was clear that online movement adaptation improved the segment separation. Since the modulated motion acts in the direction approximately orthogonal to the boundary, it introduces a tearing effect. Similar tearing motions can be observed in human executed grapefruit scooping, where a knife's reorientation lowers movement resistance by tearing through the fibers.

It should also be noted that the described task is strongly dependent on several factors. First, the nominal DMP plays an important role in the success of the task. Since the proposed method relies on torque readings gathered from the execution of the nominal trajectory, a poorly chosen movement can severely impair the medium classification. Second, the experiment is highly sensitive to the sharpness of the knife, as well as the position of the grapefruit relative to the initial pose of the knife. Nevertheless, these experiments highlight the promise of uncertainty driven cutting between mediums with differing stiffness properties. 

\section{CONCLUSIONS AND FUTURE WORK}
We present an uncertainty driven feedback control law and demonstrated its performance on the task of grapefruit segmentation. This task is selected because it resembles a common surgical procedure where a hard tumour is extracted from soft tissue, and physical material properties used to guide human surgeons. 

Automating tasks of this form is extremely challenging, as it requires cutting along an uncertain boundary, subject to visibility constraints.

Our experiments show that a simple movement correction scheme, where the movement of a robot arm is modulated along a single Cartesian axis in response to the probability of being in a given medium, significantly improves cutting performance. Future work involves the development of more complex movement adaptation schemes, and the extension of the approach to include higher level movement planning.






\bibliographystyle{IEEEtran}
\bibliography{IEEEabrv,IEEEexample,references}

\begin{thebibliography}{10}
\providecommand{\url}[1]{#1}
\csname url@rmstyle\endcsname
\providecommand{\newblock}{\relax}
\providecommand{\bibinfo}[2]{#2}
\providecommand\BIBentrySTDinterwordspacing{\spaceskip=0pt\relax}
\providecommand\BIBentryALTinterwordstretchfactor{4}
\providecommand\BIBentryALTinterwordspacing{\spaceskip=\fontdimen2\font plus
\BIBentryALTinterwordstretchfactor\fontdimen3\font minus
  \fontdimen4\font\relax}
\providecommand\BIBforeignlanguage[2]{{%
\expandafter\ifx\csname l@#1\endcsname\relax
\typeout{** WARNING: IEEEtran.bst: No hyphenation pattern has been}%
\typeout{** loaded for the language `#1'. Using the pattern for}%
\typeout{** the default language instead.}%
\else
\language=\csname l@#1\endcsname
\fi
#2}}

\bibitem{Long}
P.~Long, W.~Khalil, and P.~Martinet, ``Robotic deformable object cutting : From
  simulation to experimental validation,'' in \emph{European Workshop on
  Deformable Object Manipulation (EWDOM).}, 2014.

\bibitem{vegetables}
A.~Yamaguchi and C.~Atkeson, ``Combining finger vision and optical tactile
  sensing: Reducing and handling errors while cutting vegetables,'' in
  \emph{IEEE-RAS International Conference on Humanoid Robots}.\hskip 1em plus
  0.5em minus 0.4em\relax IEEE Computer Society, 2016, pp. 1045--1051.

\bibitem{mechcut}
X.~{Mu}, Y.~{Xue}, and Y.~{Jia}, ``Robotic cutting: Mechanics and control of
  knife motion,'' in \emph{2019 International Conference on Robotics and
  Automation (ICRA)}, May 2019, pp. 3066--3072.

\bibitem{cristini2010multiscale}
V.~Cristini and J.~Lowengrub, \emph{Multiscale Modeling of Cancer: An
  Integrated Experimental and Mathematical Modeling Approach}.\hskip 1em plus
  0.5em minus 0.4em\relax Cambridge University Press, 2010.

\bibitem{mcknight2002mr}
A.~L. McKnight, J.~L. Kugel, P.~J. Rossman, A.~Manduca, L.~C. Hartmann, and
  R.~L. Ehman, ``{MR} elastography of breast cancer: preliminary results,''
  \emph{American journal of roentgenology}, vol. 178, no.~6, pp. 1411--1417,
  2002.

\bibitem{door1}
A.~Jain, H.~Nguyen, M.~Rath, J.~Okerman, and C.~C. Kemp,
  ``\BIBforeignlanguage{eng}{The complex structure of simple devices: A survey
  of trajectories and forces that open doors and drawers},'' in
  \emph{\BIBforeignlanguage{eng}{2010 3rd IEEE RAS \& EMBS International
  Conference on Biomedical Robotics and Biomechatronics}}.\hskip 1em plus 0.5em
  minus 0.4em\relax IEEE, 2010, pp. 184--190.

\bibitem{door_grasp}
M.~{Kalakrishnan}, L.~{Righetti}, P.~{Pastor}, and S.~{Schaal}, ``Learning
  force control policies for compliant manipulation,'' in \emph{2011 IEEE/RSJ
  International Conference on Intelligent Robots and Systems}, Sep. 2011, pp.
  4639--4644.

\bibitem{door3}
A.~Jain and C.~Kemp, ``\BIBforeignlanguage{eng}{Improving robot manipulation
  with data-driven object-centric models of everyday forces},''
  \emph{\BIBforeignlanguage{eng}{Autonomous Robots}}, vol.~35, no. 2-3, pp.
  143--159, 2013.

\bibitem{Pastor}
P.~Pastor, L.~Righetti, M.~Kalakrishnan, and S.~Schaal, ``Online movement
  adaptation based on previous sensor experiences,'' in \emph{IEEE
  International Conference on Intelligent Robots and Systems}, 2011, pp.
  365--371.

\bibitem{grasp}
J.~M. {Romano}, K.~{Hsiao}, G.~{Niemeyer}, S.~{Chitta}, and K.~J.
  {Kuchenbecker}, ``Human-inspired robotic grasp control with tactile
  sensing,'' \emph{IEEE Transactions on Robotics}, vol.~27, no.~6, pp.
  1067--1079, Dec 2011.

\bibitem{obj_ident1}
A.~Schneider, J.~Sturm, C.~Stachniss, M.~Reisert, H.~Burkhardt, and W.~Burgard,
  ``\BIBforeignlanguage{eng}{Object identification with tactile sensors using
  bag-of-features},'' in \emph{\BIBforeignlanguage{eng}{2009 IEEE/RSJ
  International Conference on Intelligent Robots and Systems}}.\hskip 1em plus
  0.5em minus 0.4em\relax IEEE, 2009, pp. 243--248.

\bibitem{obj_ident2}
J.~A. Fishel and G.~E. Loeb, ``\BIBforeignlanguage{eng}{Bayesian exploration
  for intelligent identification of textures},''
  \emph{\BIBforeignlanguage{eng}{Frontiers in Neurorobotics}}, vol.~6, no.
  JUNE, 2012.

\bibitem{nonlinearcut}
F.~C. Moon and T.~Kalmár-Nagy, ``Nonlinear models for complex dynamics in
  cutting materials,'' \emph{Philosophical Transactions: Mathematical, Physical
  and Engineering Sciences}, vol. 359, no. 1781, pp. 695--711, 2001.

\bibitem{adaptcut}
G.~{Zeng} and A.~{Hemami}, ``An adaptive control strategy for robotic
  cutting,'' in \emph{Proceedings of International Conference on Robotics and
  Automation}, vol.~1, April 1997, pp. 22--27 vol.1.

\bibitem{salami}
A.~G. Atkins, X.~Xu, and G.~Jeronimidis, ``Cutting, by `pressing and slicing,'
  of thin floppy slices of materials illustrated by experiments on cheddar
  cheese and salami,'' \emph{Journal of Materials Science}, vol.~39, no.~8, pp.
  2761--2766, Apr 2004.

\bibitem{DeepMPC}
I.~Lenz, R.~Knepper, and A.~Saxena, ``Deep{MPC}: Learning deep latent features
  for model predictive control,'' in \emph{Robotics: Science and Systems XI},
  2015.

\bibitem{tactileMPC}
S.~Tian, F.~Ebert, D.~Jayaraman, M.~Mudigonda, C.~Finn, R.~Calandra, and
  S.~Levine, ``Manipulation by feel: Touch-based control with deep predictive
  models,'' in \emph{2019 International Conference on Robotics and Automation
  (ICRA)}, 2019.

\bibitem{pr2salad}
M.~C. {Gemici} and A.~{Saxena}, ``Learning haptic representation for
  manipulating deformable food objects,'' in \emph{2014 IEEE/RSJ International
  Conference on Intelligent Robots and Systems}, Sep. 2014, pp. 638--645.

\bibitem{Ijspeert}
A.~J. {Ijspeert}, J.~{Nakanishi}, H.~{Hoffmann}, P.~{Pastor}, and S.~{Schaal},
  ``Dynamical movement primitives: Learning attractor models for motor
  behaviors,'' \emph{Neural Computation}, vol.~25, no.~2, pp. 328--373, Feb
  2013.

\bibitem{DMPobstacle}
P.~{Pastor}, H.~{Hoffmann}, T.~{Asfour}, and S.~{Schaal}, ``Learning and
  generalization of motor skills by learning from demonstration,'' in
  \emph{2009 IEEE International Conference on Robotics and Automation}, May
  2009, pp. 763--768.

\bibitem{outcome}
P.~{Pastor}, M.~{Kalakrishnan}, S.~{Chitta}, E.~{Theodorou}, and S.~{Schaal},
  ``Skill learning and task outcome prediction for manipulation,'' in
  \emph{2011 IEEE International Conference on Robotics and Automation}, May
  2011, pp. 3828--3834.

\bibitem{lwpr}
S.~Vijayakumar and S.~Schaal, ``Locally weighted projection regression: An o(n)
  algorithm for incremental real time learning in high dimensional space,''
  \emph{Proceedings of the Seventeenth International Conference on Machine
  Learning (ICML 2000)}, vol. Vol. 1, 05 2000.

\end{thebibliography}

\end{document}